\documentclass[runningheads]{llncs}
\usepackage[title,toc,titletoc,header]{appendix}
\usepackage{subfig}
\usepackage{mwe}
\usepackage{graphicx}
\usepackage{amsmath,amssymb} 
\usepackage{multirow}
\usepackage{hyperref}
\usepackage{color}
\PassOptionsToPackage{hyphens}{url}\usepackage{hyperref}
\graphicspath{{img/}}
\usepackage[width=122mm,left=12mm,paperwidth=146mm,height=193mm,top=12mm,paperheight=217mm]
{geometry}
\expandafter\def\expandafter\UrlBreaks\expandafter{\UrlBreaks
  \do\a\do\b\do\c\do\d\do\e\do\f\do\g\do\h\do\i\do\j%
  \do\k\do\l\do\m\do\n\do\o\do\p\do\q\do\r\do\s\do\t%
  \do\u\do\v\do\w\do\x\do\y\do\z\do\A\do\B\do\C\do\D%
  \do\E\do\F\do\G\do\H\do\I\do\J\do\K\do\L\do\M\do\N%
  \do\O\do\P\do\Q\do\R\do\S\do\T\do\U\do\V\do\W\do\X%
  \do\Y\do\Z}
\begin{document}

\pagestyle{headings}
\mainmatter

\title{Report: Dynamic Eye Movement Matching and Visualization Tool in Neuro Gesture} 

\titlerunning{Dynamic Eye Movement Matching and Visualization Tool in Neuro Gesture}

\author{Qiangeng Xu , John R. Kender}

\institute{Computer Science Department,\\
	Columbia University\\
	\email{qx2128@columbia.edu, jrk@cs.columbia.edu}
}

\maketitle
\begin{abstract}
In the research of the impact of gestures using by a lecturer, one challenging task is to infer the attention of a group of audiences. Two important measurements that can help infer the level of attention are eye movement data and Electroencephalography (EEG) data. Under the fundamental assumption that a group of people would look at the same place if they all pay attention at the same time, we apply a method, "Time Warp Edit Distance", to calculate the similarity of their eye movement trajectories. Moreover, we also cluster eye movement pattern of audiences based on these pair-wised similarity metrics. Besides, since we don't have a direct metric for the "attention" ground truth, a visual assessment would be beneficial to evaluate the gesture-attention relationship. Thus we also implement a visualization tool.

\end{abstract}

\section{Introduction}

It is well studied that people's gazing position would reveal the visual stimuli that people receive. The saliency of an item, defined by the state or quality by which it stands out relative to its neighbors is considered to be a key attention mechanism that facilitates in many learning tasks~\cite{chen2016sca,xu2017generative}. Several recent neural science studies have discovered specific tissues and neurons that would support such mechanism. For example, Snow et al.\cite{snow2009impaired} studied pulvinar nuclei which modulates physical/perceptual salience in attentional selection. Besides, Baliki et al.\cite{baliki2013parceling} discovered D1-type medium spiny neurons and D2-type medium spiny neurons within the NAcc shell assigns aversive motivational salience to aversive stimuli. 

Several previous studies concentrated on the relation between viewer's attention and visual stimuli such as Loschky et al.\cite{loschky2015would}, Dewhurst et al.\cite{dewhurst2012depends} and Valuch et al.\cite{valuch2015influence}. They used eye tracking data in order to discover the similar eye trajectories across viewers. More interestingly, the study introduced by Burleson-Lesser et al.\cite{burleson2017collective} has revealed that video material also synchronises judgments and the similarity of eye movement could help predict viewer's preference. 

Thus, under the assumption that higher similarity of eye movement indicates higher attention of a group of audience, we try to use eye tracking data to find the similarity of eye movement and cluster the audiences into subgroups based on their eye movement similarities. We believe such effort would help us to find the more "attractive moment" in a lecture video that would get higher attention from audiences, indicated by more "homogeneous" eye movement clustering result. On the other hand, we would also expect such clustering could help us to find some outlier among audiences(e.g. some people don't pay attention to anything). 

Another work we have done is to re-design the visualization tool previously implemented in Visual Studio. To cope with the diverse working environment that our team members have, we conduct frame-work selection between a few popular candidates. We select Qt in the end and added EEG graph into the visualization tool. The contributions of our work to the Neuro Gesture project are two folds:
\begin{quote}
	\begin{description}

  \item[$\bullet$] We apply Time Warp Edit Distance methods to calculate the pair-wise similarity between each eye movement trajectories pair. Then we use the similarity matrix to cluster the audience eye movement pattern by using relevant communities clustering method introduced by Le Martelot et al.\cite{le2013fast}
  \item[$\bullet$] We implement a visualization tool. The main functionalities include: 1. eye fixation and trails superimposing on the video and 2. 64 channels of EEG graph on sync with the video lapse. After considered several framework, we select Qt, the most compatible cross-platform SDK.
 \end{description}
\end{quote} 
The rest of the report is organized as follows. We introduce the similarity matrix and clustering method in Section 2 and visualization tool in section 3. In section 2, the related work and model selection would be discussed in 2.1, the detailed algorithm of similarity matrix would be shown in 2.2 and the parameter tuning would be shown in 2.3. The clustering algorithm would be shown in 2.4. The study of correlation between similarity matrix and question answer correctness would be shown in 2.5. In section 3, we introduce the visualization tool's design in 3.1 and illustrate the display in 3.2. In the end, we summarize our work and future challenges in section 4.
	  
\section{Eye Movement Similarity Matching and Clustering}

The form of eye-tracking data is essentially multi-dimensional time-series data of multiple objects. Thus here we discuss several time-series matching methods and their strength and weakness if applied our task.

\subsection{Related Works and Model Selection}
\subsubsection{Inter-subject correlation of eye movement} is an emerging topic which would lead toward a better understanding of correlated activity between individuals under stimuli. Some previous studies such as Burleson-Lesser et al.\cite{burleson2017collective} employed tools from statistical mechanics which is very powerful to explain emergent properties of the local interactions between subjects. They have demonstrated that the audiences' eye movement would be distributed under a balance between randomness and alignment. They modeled the distribution of direction as it is entirely random yet the observed mean \(\mu_i\) and pair-wise correlations \(C_{ij}\) could be derived. The model is also studied by Mora et al.\cite{mora2011biological}:
\begin{align*}
	p(\overrightarrow{\sigma}) \propto exp(-\sum_{i}\vec{h_i}\cdot\overrightarrow{\sigma_i} - \sum_{i,j > i}J_{ij}\overrightarrow{\sigma_i}\cdot\overrightarrow{\sigma_j})\
\end{align*}
Adopting same measurement in the statistic mechanics studies such as Cavagna et al.\cite{cavagna2010scale}, Bialek et al.\cite{bialek2012statistical} and Tka{\v{c}}ik et al.\cite{tkavcik2015thermodynamics}, they computed the correlation among eye movement using direction of every pair of two viewers:
\begin{align*}
	C_{ij} = \langle\overrightarrow{\sigma_i}(t)\cdot\overrightarrow{\sigma_j}(t)\rangle_t - \overrightarrow{\mu_i}\cdot\overrightarrow{\mu_j} \\
	where: \overrightarrow{\mu_i} = \langle\overrightarrow{\sigma_i}(t)\rangle_t
\end{align*}
However, considering our setting of the experiment(we let each viewer look at the lecture video in a individual room), the homophily of eye movement could only be explained by the video content and the attention of an individual at a certain moment. The "interaction" between audiences doesn't have a physical meaning but only serves as an analogy. More importantly, on purpose of inferring the attention of audience to visual stimuli, we choose to rely on fixation instead of eye movement direction. Since We have discovered many cases that would cause viewer having different eye movement direction while actually getting attracted by the same subject. For example, when the camera is shifting and the scene is switching, people's eye gazing position would be random at first, then, concentrated on the same object that camera is switching to. The movement direction during this period would be very diverse, which hinders the fact that viewers actually have similar "attention". Besides, the method used in \cite{burleson2017collective} would provide little tolerance to mismatch on temporal dimension. From our observation, individual viewers always have different response time to visual stimuli. Many medical studies such as Jain et al.\cite{jain2015comparative} has discovered such variation between individuals. Thus we explore the methods in time series matching by using fixation data to serve our research.

\subsubsection{Time series similarity measures} 
\begin{figure}[!ht]
   \centering
   \subfloat[][]{\includegraphics[width=.47\textwidth]{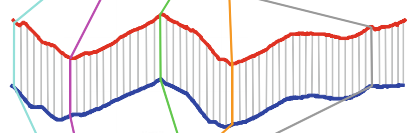}}\quad
   \subfloat[][]{\includegraphics[width=.47\textwidth]{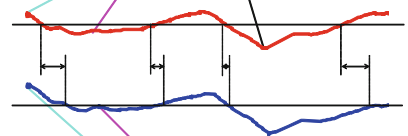}}\\
   \subfloat[][]{\includegraphics[width=.47\textwidth]{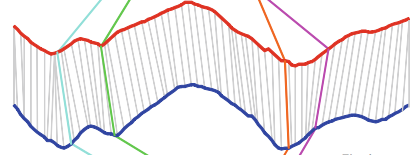}}\quad
   \subfloat[][]{\includegraphics[width=.47\textwidth]{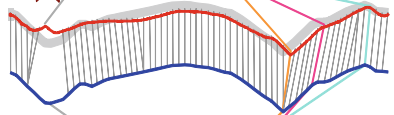}}
   \caption{The distance measure is proportional to the length of the gray lines. Method shown in (a) is a \textit{lock step} measure. The "one to one" mapping is enforced. The method illustrated in (b) is a \textit{Threshold} measure. In (c), an Elastic measure such as DTW is shown. The method would allowed "one-to-many" mapping of the data points, but each data points must be matched. In (d), the methods such as TWED is illustrated. Additional to "one-to-many" mapping of the data points, it also allows the possibility of not matching  points. }
   \label{fig:dist}
\end{figure}

is a central task in the domain of analysis, prediction or classification of information unfolding over time. Given time series \(T_1\) and \(T_2\), a similarity function calculate one point of \(T_1\) and one corresponding point of \(T_2\) having the same timestamp belongs to the family of \textit{Lock-Step Measure}. The lock-step measures including Euclidean Distance matching introduced by Faloutsos et al.\cite{faloutsos1994fast} along with its variants using \(L_p\)-norms(including Manhattan norm, Maximum norms, etc.) introduced by Yi et al.\cite{yi2000fast}. The advantage of these methods is their simplicity. The matching map is shown in \footnote{Figure 1 is adapted from \cite{wang2013experimental}} Figure~\ref{fig:dist}(a). However, in the case of eye-tracking data, because of the time  inaccuracy in eye fixation measurement and different response time among individuals, these methods would give us large distance between an audience having faster response and an audience who is relatively slower, even both their trajectories are following the same pattern. 

Another rather novel family of approaches is \textit{Threshold-Based Measure}. For example, the TQuEST distance introduced by A{\ss}falg et al.\cite{assfalg2006similarity} is using a threshold parameter \(\tau\) and transform a time series to threshold-crossing time intervals. Each time interval is treated as a point in two dimensional space. the Minkowski sum of the two sequences served as the similarity. The matching map is shown in Figure~\ref{fig:dist}(b). In general this approach would be more suitable to data that has similar amplitude value(fixation of eye gazing point) but different temporal value. However, the eye-tracking data we collect shows diverse pattern both in temporal value and amplitude value.

Other families of approaches also have been well studied. The \textit{Feature-Based Measures} such as Fourier coefficients matching introduced by Oppenheim et al.\cite{oppenheim1999discrete} used discrete Fourier transforms of the raw time series to filter out high-frequency coefficients, making it a efficient methods to remove rapidly fluctuating signal components. Yet, in our case, it also suffered from not being able to adapt to the differences among individual eye responses. The family of \textit{Pattern-Based Measure} including SpADe introduced by Chen et al.\cite{chen2016sca} matches segments within entire time series by adjusting(shifting or scaling) value in both temporal and amplitude dimensions. 
Although this method concentrates on matching patterns between time series, in our case, shifting the amplitude value(e.g. fixation) would be problematic. Two audiences would generally pay attention to same object(e.g. the lecturer or picture in slides), thus similar fixation, if they are both attracted by the lecture. 

The most effective family of matching methods for our data might be \textit{Elastic Measure}. Methods in this family generally would allowed matching between one-to-many and one-to-none points. In our case, some of the fixation data points are missing and some people's attention would stay relatively longer in an object or move slower than others. The most popular method in this family would be dynamic time warping(DTW) introduced by Berndt et al.\cite{berndt1994using} The method aligns the time series in the temporal domain that minimize the cost of the matching distance after the alignment. Dynamic programming could be applied recursively to solve this problem. 
\begin{align*}
	Minimize: d_{DTW(X,Y)} = D_{M,N} \\ 
	D_{i,j} = f(x_i,y_j)  + Min\left\{
                \begin{array}{ll}
                  D_{i,j-1} \\
                  D_{i-1,j} \\ 
                  D_{i-1,j-1} \\
                \end{array}
              \right. \\
	while: i= 1,...,M, j = 1,...,N 
\end{align*}
In the study of Wang et al.\cite{wang2013experimental},Very few measures have been proposed that systematically outperform DTW for a number of different data sources. The matching map is shown in Figure~\ref{fig:dist}(c). However, the DTW method would entirely ignore the time lag between eye trails, which might also be an unwanted behavior. Instead, we study the edit distance time warping methods such as Edit distance on real sequence introduced in \cite{chen2005robust}, Edit
Distance with Real Penalty (ERP) introduced in \cite{chen2004marriage} and Time Warp Edit Distance(TWED) introduced by Marteau et al.\cite{marteau2009time}. Among those extensions, TWED is the most suitable one to the eye-tracking data since it incorporates both time stamp differences and editable series matching. TWED includes a deletion penalty \(\lambda\) and stiffness parameter \(\gamma\) to cope with three different conditions each step. The whole objective could be summarized as follows:
\begin{align*}
	Minimize &: d_{TWED}(x,y) = \delta_{M,N} \\
    having &: 
\end{align*}
\begin{align*}
	\delta_{\lambda,\gamma}(A_1^p, B_1^q) &= Min\left\{
                \begin{array}{ll}
                  \delta_{\lambda,\gamma}(A_1^{p-1}, B_1^q) + d(a_p, a_{p-1}) + \gamma d(t_{a_p}, t_{a_{p-1}}) + \lambda \hspace{0.5cm} deleteA \\
                  \delta_{\lambda,\gamma}(A_1^{p-1}, B_1^{q-1}) + d(a_p, b_q) + d(a_{p-1}, a_{p-1}) + \\ \hspace{2.5cm} \gamma d(t_{a_p},t_{b_q}) + \gamma d(t_{a_{p-1}},t_{b_{q-1}}) \hspace{1cm} match \\
                  \delta_{\lambda,\gamma}(A_1^p, B_1^{q-1}) + d(b_{q-1}, b_q) + \gamma d(t_{b_q}, t_{b_{q-1}}) + \lambda \hspace{0.5cm} deleteB \\
                \end{array}
              \right.
\end{align*}
The matching map of TWED is shown in Figure~\ref{fig:dist}(d).
By looking at the form of TWED, we can easily find the DTW, ERP, LCSS methods are all TWED's special case. We will illustrate the detail of TWED algorithm applied to eye-tracking data in the next section.

\subsubsection{Clustering based on pair-wise relationship between data points} is a well studied problem with various solutions. However, since the number of group or the granularity of the cluster is unknown, adopting traditional clustering method would force us to determine this number. However, to seek the optimal number of clusters might be an unwanted operation since we want to make the clustering criteria consistent through whole lecture(we separate video to several sub sections to do the clustering independently). Some part of the lecture might simply be more attractive and people would pay more attention to similar object, other parts may not be delivering interesting information thus people starting looking around. Under certain scenario, the cluster number should not be the same but vary accordingly. Thus we adopt community detection to solve this problem. 

Community detection, compared to traditional clustering method, the number of cluster(community) is always unknown. Most of the methods rank communities by using a criterion. Also studied by Simon et al.\cite{simon1991architecture}, a dataset often have several levels of granularity resulting various clustering at several resolutions. Small clusters could be generated by fine scale analysis but a large scale analysis could result in larger clusters. Many studies including Le Martelot et al.\cite{le2013multi} and Huang et al.\cite{huang2011towards} have proposed criteria designated to suffice multi-scale analysis.  However, they are limited in their efficiency and accuracy. In stead, we find the method proposed by Le Martelot et al.\cite{le2013multi} could enable fast multi-scale community detection on large networks with global and local criteria. Their method have also been effectively adopted in eye-tracking clustering by Burleson-Lesser et al.\cite{burleson2017collective}. The detail implementation and result would be introduced in following sections.

\subsection{Method and implementation of eye movement similarity matrix}
Our project code Eye-movement-similarity-clustering are released at\\
\url{https://github.com/Xharlie/Eye-movement-similarity-clustering}
\subsubsection{Data Preprocessing}
:\\
Our data come from the proprietary software which runs the eye-tracking machine while recording the audience eye fixation. The output file information and format are illustrated in \textbf{Appendix A}. We adopt the preprocessing implementation in Burleson-Lesser et al.\cite{burleson2017collective}. We filter the fixation data with 80 millisecond triangular window. Participants would be excluded from the video if they had over 20$\%$ missing data in a video. All missing data would be set to 0. A sparse principal component analysis was run on each dimension. The operation inserted a linear interpolation over viewers to missing samples. We also divide the data for a video into 30 second data files. The sample file is shown in the project's data folder. Each file represent x or y position for an object in a 30 seconds session.
\subsubsection{Similarity Matrix} 
\begin{figure}
\centering
\includegraphics[height=6.5cm]{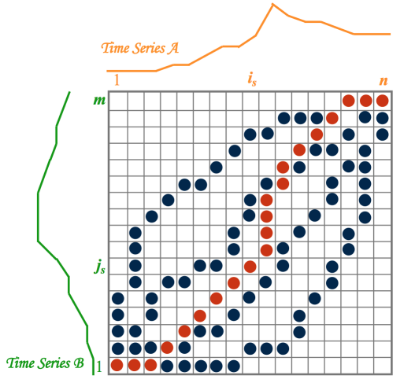}
\caption{the path contains lowest distance between series A and B would be selected as the result.}
\label{fig:TWED_Dynamic_Programming}
\end{figure}

:\\
The frame rate of the eye tracking data could be very high(capable to be more than 1000 frames per second). Even after preprocessing, we still have eye-tracking frame rate around 120 per second. However, since the video's(we use TED talk as an example) frame rate are only 32 frames per second. Thus first of all, we down-sample the eye-tracking data to the same frame rate as the video.
After that, we will calculate the similarity matrix by using TWED. TWED has two hyper parameters: deletion penalty \(\lambda\) and stiffness parameter \(\gamma\) in order to cope with three different conditions each step. The "deletion" penalty or according to Serra et al.\cite{serra2014empirical} the "mismatch" penalty is a constant cost setting to compensate the situation whenever we decide one of the series would stay at the same point while the other one would step forward. The stiffness parameter would add a cost proportional to the distance between two series temporal distance at a certain step. In the original literature of the algorithm in \cite{marteau2009time}, the "elastic cost" has the form $\gamma d(t_{a_{p-1}},t_{b_{q-1}})$. Here, since the distance between $t_{a_{p}}$ and $t_{b_{q}}$ or $t_{a_{p-1}}$ and $t_{a_{p}}$ are proportional to the difference of index no matter what time unit we choose, we can just write the distance of time as the distance of frame indices so that the time unit could be absorbed into $\gamma$. Now, the simpler equation for each step is:
\begin{align*}
	\delta_{\lambda,\gamma}(A_1^p, B_1^q) &= Min\left\{
                \begin{array}{ll}
                  \delta_{\lambda,\gamma}(A_1^{p-1}, B_1^q) + d(a_p, a_{p-1}) + \gamma  + \lambda \hspace{0.5cm} deleteA \\
                  \delta_{\lambda,\gamma}(A_1^{p-1}, B_1^{q-1}) + d(a_p, b_q) + d(a_{p-1}, a_{p-1})  \\ \hspace{2.5cm} + 2\gamma|p-q| \hspace{2cm} match \\
                  \delta_{\lambda,\gamma}(A_1^p, B_1^{q-1}) + d(b_{q-1}, b_q) + \gamma + \lambda \hspace{0.5cm} deleteB \\
                \end{array}
              \right.
\end{align*}
Here we just fill in all the possible steps from bottom left corner to upper right corner(see Fig.\ref{fig:TWED_Dynamic_Programming}). Each element contains the minimum possible cost to reach that element from the origin. 
The algorithm of TWED in our application is listed in \textbf{Appendix B} The implementation is inside the TWED.py file in our project repository. \\
After we enumerate through all combination of eye trail pairs, we would get a N by N distance matrix in which each cell $C_{i,j}$ contains the TWED result for eye trail i and j.
To transform the distance matrix to similarity matrix, we simply normalize each element by dividing the largest value among all the elements in the matrix and subtracting it by 1. Thus, the similarity matrix we get for a video in a time window would contains all pair-wise similarity values between 0 to 1. The matrix is a symmetric matrix with all value 1 on its diagonal(since the ith eye trail would have 100\% similarity with itself). 

\subsubsection{Parameter Selection of TWED} 
\begin{figure}[!ht]
   \centering
   \subfloat[][]{\includegraphics[width=.99\textwidth]{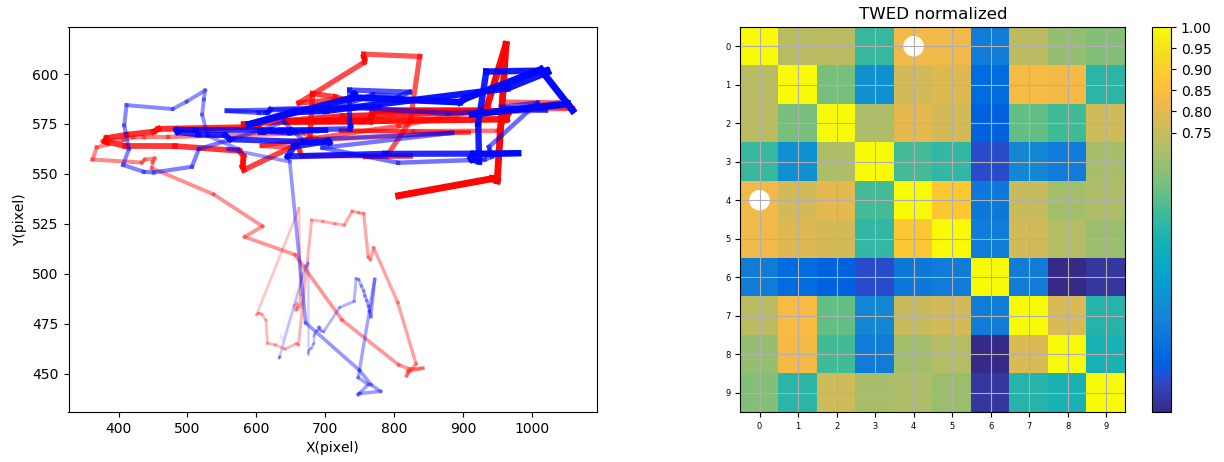}}\\
   \subfloat[][]{\includegraphics[width=.99\textwidth]{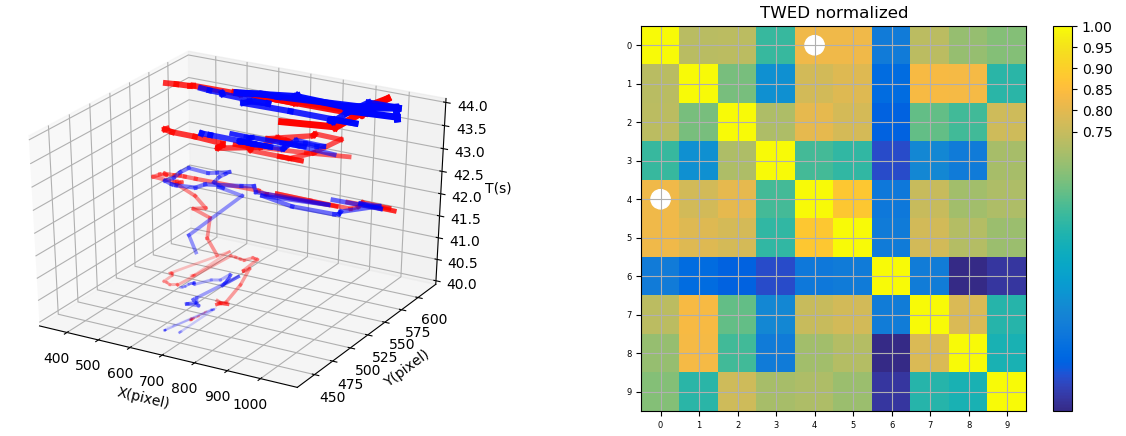}}\\
   \caption{These are the samples of visualization and similarity matrix. We select the time window from the 33 second to the 38 second in the TED Talk Carol. The similarity matrix is calculated when setting $\lambda = 5000,\gamma = 5000$. We pick object 4 and object 0 and marked the similarity value in the matrix by white dots. (a) shows the 2D trail visualization. Two trails have different colors and the colors are proportionally getting stronger along temporal dimension. (d) shows the 3D trail visualization. The colors are also proportionally getting stronger along temporal dimension. The time dimension is explicitly shown as T axis.}
   \label{fig:TWED parameter}
\end{figure}

:\\
The hyper-parameter in TWED algorithm $\lambda,\gamma$ would affect the importance of temporal step matching. With $\lambda,\gamma$ both set to zero, the TWED algorithm would automatically select the matching between two time series with minimum distance. However, in our case, such matching strategy would be problematic if we match two viewer's gazing fixations at an object's position before and after a shifting of that object in the video scene. Here we will exam the impact of the parameter selection by visual evaluation. We pick 6 $\times$ 6 combination of $\lambda,\gamma$ ($\lambda$ and $\gamma$ iterate with values [0, 1000, 5000, 10000, 20000]). Then we pick several window of 5 second for the TED talk Carol and visualize the movement of different pairs and their values under different hyper-parameter selections. We implement both 3d and 2d visualization of 2 eye movement trails, a sample visualization is shown in Fig. \ref{fig:TWED parameter}. We also select parula color map to get a better illustration of the similarity matrix. We would like to get some quantitative ground truth to facilitate the parameter selection, but seems the visual assessment is the only feasible method for now.
So far, depends on visual assessment, we would select $\lambda$ as 5000 and $\gamma$ as 5000. More comparisons between different parameter settings could be found in \textbf{Appendix C}. After selecting the optimal parameters, we set the optimal parameters and run through all the video segments with time window of 5 seconds as well as 30 seconds. We then, save the result under the "result" folder in our project repository.

\subsubsection{Eye Movement Clustering}
\begin{figure}[!ht]
   \centering
   \subfloat[][]{\includegraphics[width=.47\textwidth]{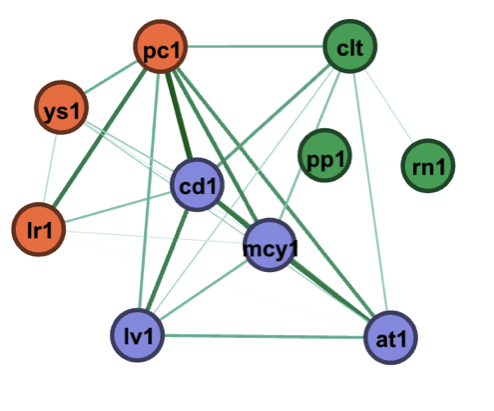}}\quad
   \subfloat[][]{\includegraphics[width=.47\textwidth]{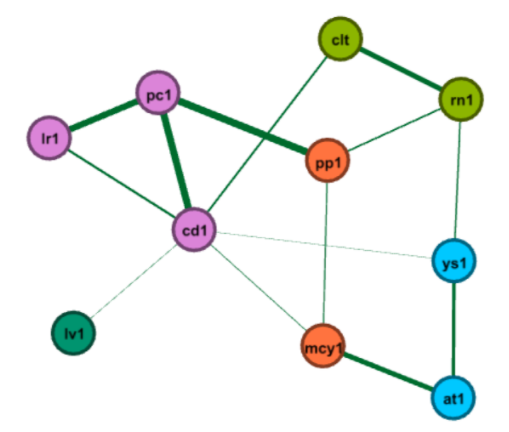}}\\
   \subfloat[][]{\includegraphics[width=.47\textwidth]{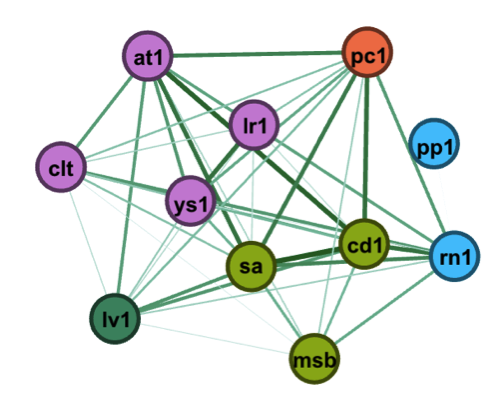}}\quad
   \subfloat[][]{\includegraphics[width=.47\textwidth]{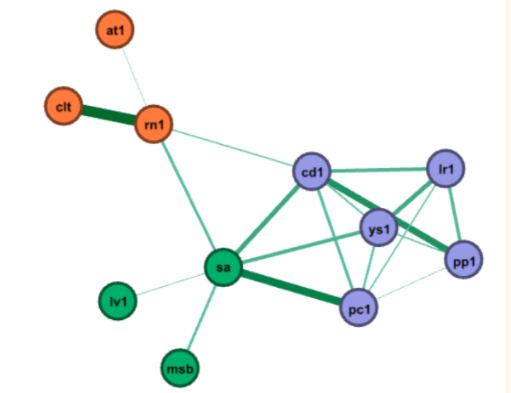}}\\
   \caption{(a) is the clustering result of 200s to 230s in Carol using TWED similarity matrix (b) is the clustering result of 200s to 230s in Carol using adjacent matrix introduced in \cite{burleson2017collective} (c) is the clustering result of 290s to 320s in Carol using TWED similarity matrix (d) is the clustering result of 290s to 320s in Carol using adjacent matrix introduced in \cite{burleson2017collective}}
   \label{fig:clustering result}
\end{figure}

We adopt most of the clustering code in \cite{burleson2017collective}. Besides replacing the $C_{ij}$ vector direction adjacenct matrix by our similarity matrix, we also directly use the $C_{ij}$ as the $J_{ij}$ in \cite{burleson2017collective}. To better visualize the clustering, we also map the object's index in TWED result file to the object's name. The algorithm is shown in \textbf{Appendix D} and the matlab code is located in the "clustering" folder in our project repository.
After saving the clustering result into the location:
\texttt{clustering}$\backslash$\texttt{clustering\char`_result}, we also use the Gephi to visualize the clustering result. We select a 30 second window staring the 290 second and a 30s window starting the 290s in TED Talk Carol video. The clustering result is shown at Fig.\ref{fig:clustering result}
\subsubsection{Correlation Between Similarity Matrix and Question Answer Correctness}
\begin{figure}
\centering
\includegraphics[height=6.5cm]{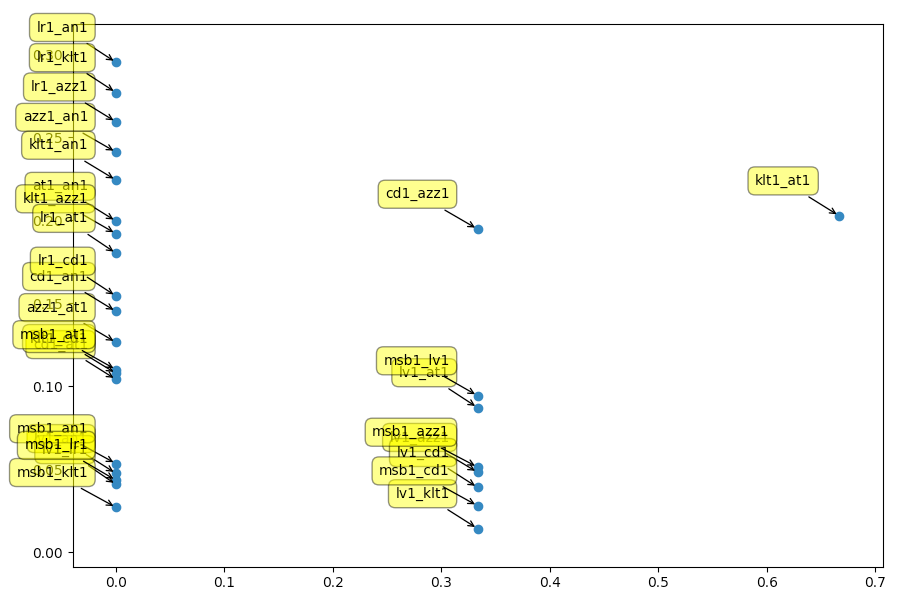}
\caption{The x axis represents the question answer similarity. The y axis shows the similarity value of the eye movements}
\label{fig:QAandSimilarity}
\end{figure}

We also examine the relationship between the TWED similarity matrix and question answer correctness. \\
Because of lacking ground truth of "attention", we explore the possibility to use question  answer correctness as the evaluation standard. We select a 30 seconds window(the 405s to the 435s) in TED talk Simon video in which the content is related to 5 questions(s3C, s1B, s1E, s3F, s3G). The timing of these questions could be found at \url{https://docs.google.com/spreadsheets/d/CCMoVJfCyiCMZxf0D1mexE4uELeodQPhRrfBnikXDtU/edit} and the correctness of these questions could be found at \url{https://docs.google.com/spreadsheets/d/1bDehGoP68VTa7ycggp7G9yOi2FHYNyl6S51Ne4Rolzc/edit} \\
Under the assumption that more similar the viewers answering the questions, more likely they pay the same level of attention to the lecture. We calculate a N by N answer distance penalty matrix holding the answer difference penalty between each pair of viewers. If both viewers have correct answer to one question, we would add 1 to their distance penalty value. If both or one of the viewers have incorrect answer, we don't add value to their distance penalty. The higher the distance penalty, the more similar the two viewers' attention is in this time window. We then, along with the similarity matrix of this time window, plot the value of answer distance penalty as x value and similarity as y value. The result is showing in Fig. \ref{fig:QAandSimilarity}

As we can see, since the number of questions is too small, it is relatively hard to apply a regression to reveal the correlation. However, by only looking at this plot, we can see in this period of time, the similarities of eye movement between viewers are relatively low and the similarities of their answers are low at the same time.
  
\section{Eye-Tracking Overlay and EEG Visualization Tool}
The code of the visualization tool could be found at \url{https://github.com/hectorcho/visproj} The newest developed version is inside its \texttt{qt5\char`_structuralize} folder.
\subsection{Design and Framework}
\begin{figure}
\centering
\includegraphics[height=6.5cm]{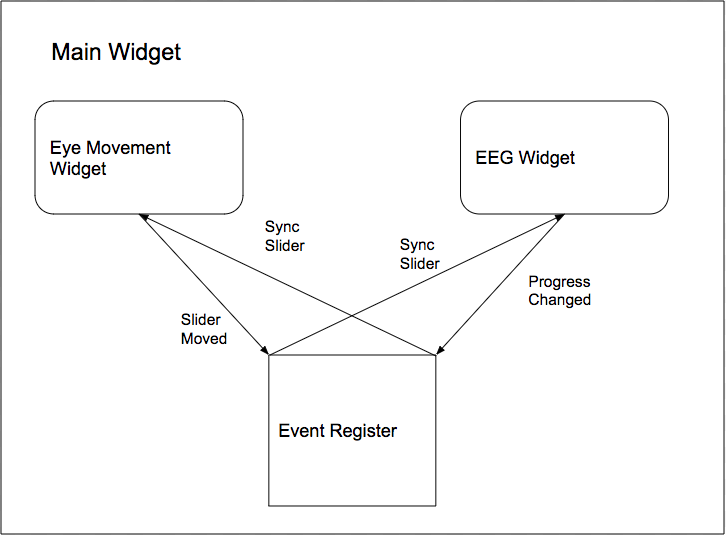}
\caption{The events include: syncSlider, pauseVideo, startVideo, loadVideo, loadEyeTracking, loadEEG, etc.}
\label{fig:widget}
\end{figure}

Due to compatibility issue across different development environment, we decide to implement a visualization tool by using PyQt5. QT is basically the most flexible framework for cross-platform GUI development. In addition to PyQt5, we also rely on pyqtgraph, a library help plot dynamic data. The detail of dependencies is shown in \textbf{Appendix E}. \\
The requirement of our visualization tool includes a video player, a eye fixation canvas super-imposed above the video player and a EEG data graph that can dynamically update EEG data for all channels. \\
An important concept of Qt GUI is widget, which is a complex development unit that contains various basic functionalities. After a few structure changes, we customize our own widget by inheriting the most basic widget and adding corresponding functions. The tool includes three components: 1.a eye movement and video playing widget; 2.an EEG widget; 3.a main window widget that in charge of the GUI layout and the inter-widget communication. We apply observer-listener pattern to register several events to make the eye movement widget and EEG widget on sync. A widget level design is shown in Fig. \ref{fig:widget}

To include with the functionality of eye movement super-imposing, we choose not to create a separate widget but adding a graph video element as the video player's mirror inside a graph view and dynamically add and delete graphic elements such as dots and lines to illustrate the eye movement. The "open Eye" button will read in several eye tracking files, preprocess the fixation data and generate a dictionary. The dictionary holds every object's eye gazing dot, the eye trail lines of previous steps for each eye movement record and a designated color to draw these elements. Whenever the slider of the video moved, the drawing will be triggered to add, delete, move or transform those graphic elements(gazing dots, trail lines). Moreover, the gazing dots would change their size depends on the duration of gazing at each record. \\
By selecting audiences in the comboBox, the corresponding eye movement graphic elements would be added to or removed from the graphic view. The position of the fixation has been normalized according to the original screen's resolution and the video player's resolution.

We use the pyqtgraph library to develop the EEG graph widget. The EEG data contains 64 channels, but we allow user to select any numbers of channels to be shown in the canvas. Their plots would be added or removed accordingly. If the video lapse slider change its position, the EEG graphic would also be triggered to update the corresponding EEG graph. We choose to display all EEG values between 5 seconds before and after the current video time. The detail of tool's functions is shown in \textbf{Appendix F} \\

\subsection{The Display of the Visualization Tool}

\begin{figure}[!ht]
   \centering
   \subfloat[][]{\includegraphics[width=.47\textwidth]{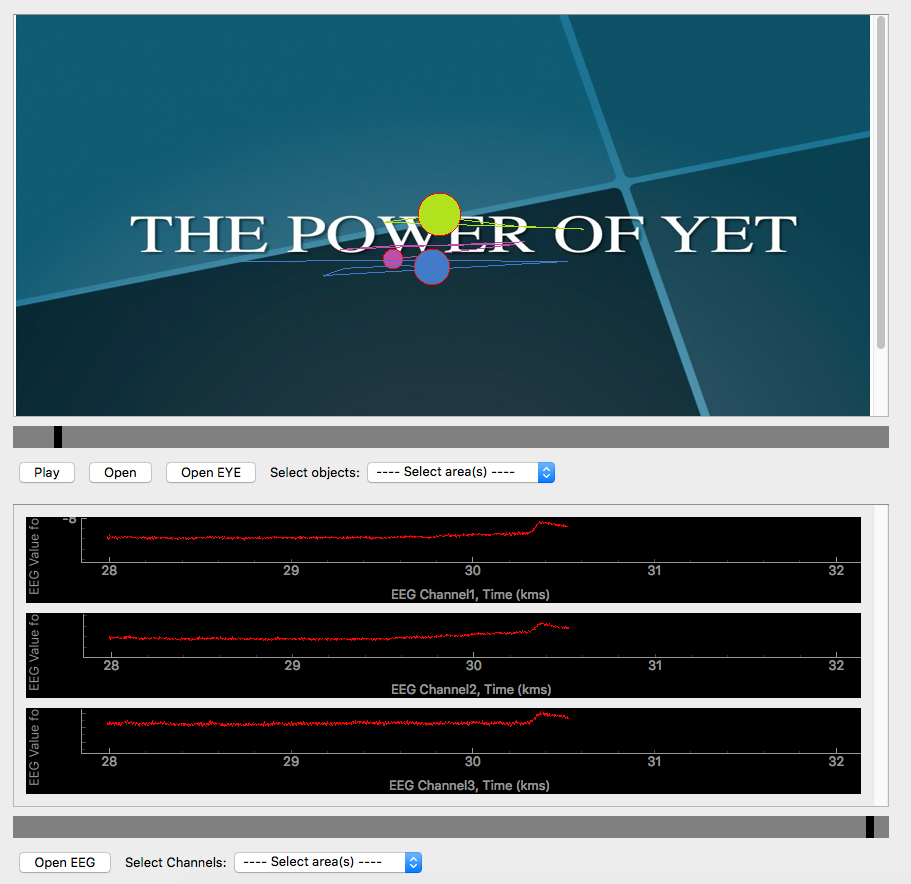}}\quad
   \subfloat[][]{\includegraphics[width=.47\textwidth]{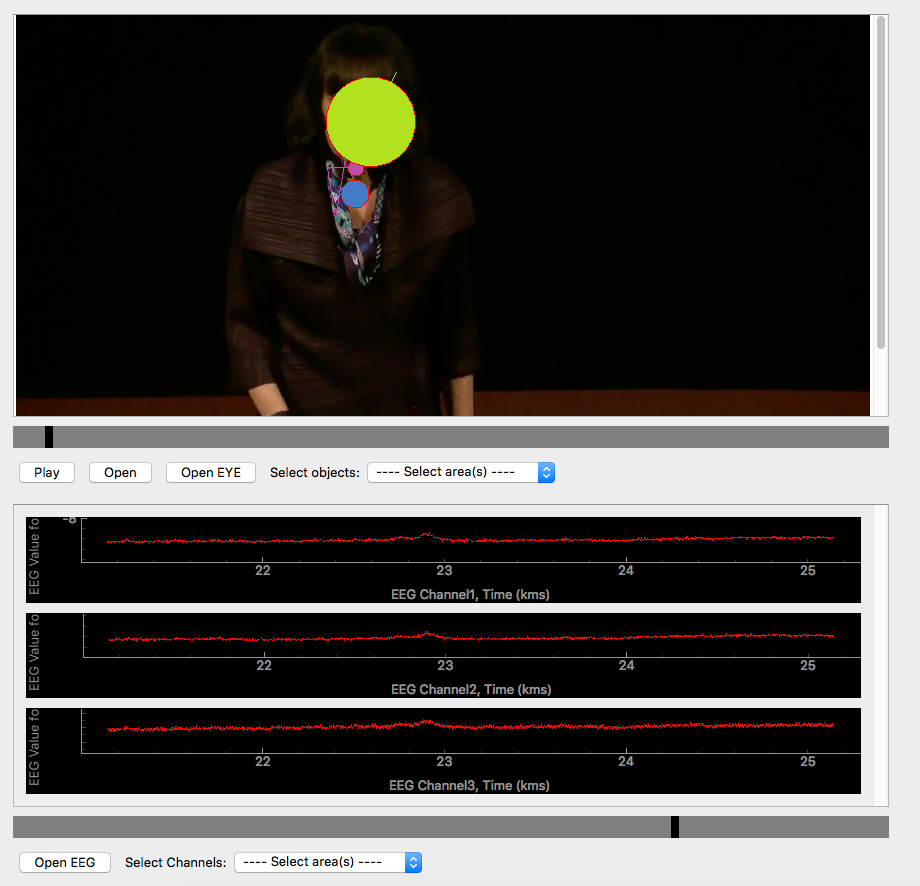}}\\
   \caption{(a) is the display when viewers are busy reading the slides. (b) is the display when viewers are only gazing at the speaker}
   \label{fig:toolnormal}
\end{figure}

\begin{figure}[!ht]
   \centering
   \subfloat[][]{\includegraphics[width=.47\textwidth]{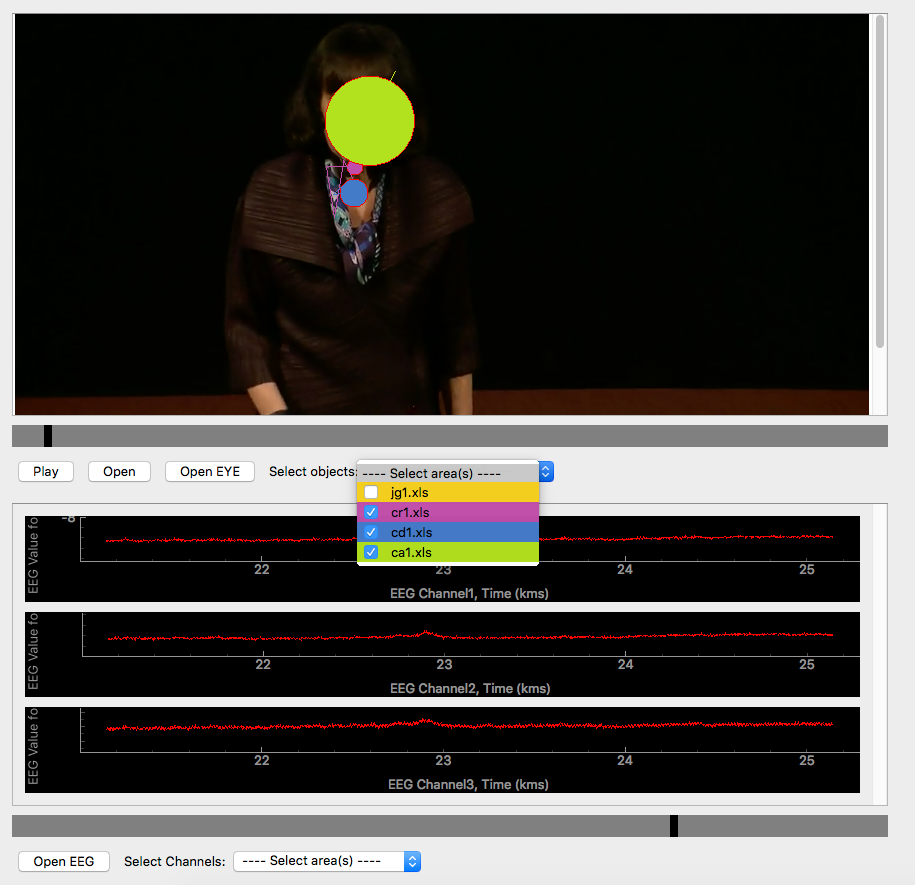}}\quad
   \subfloat[][]{\includegraphics[width=.47\textwidth]{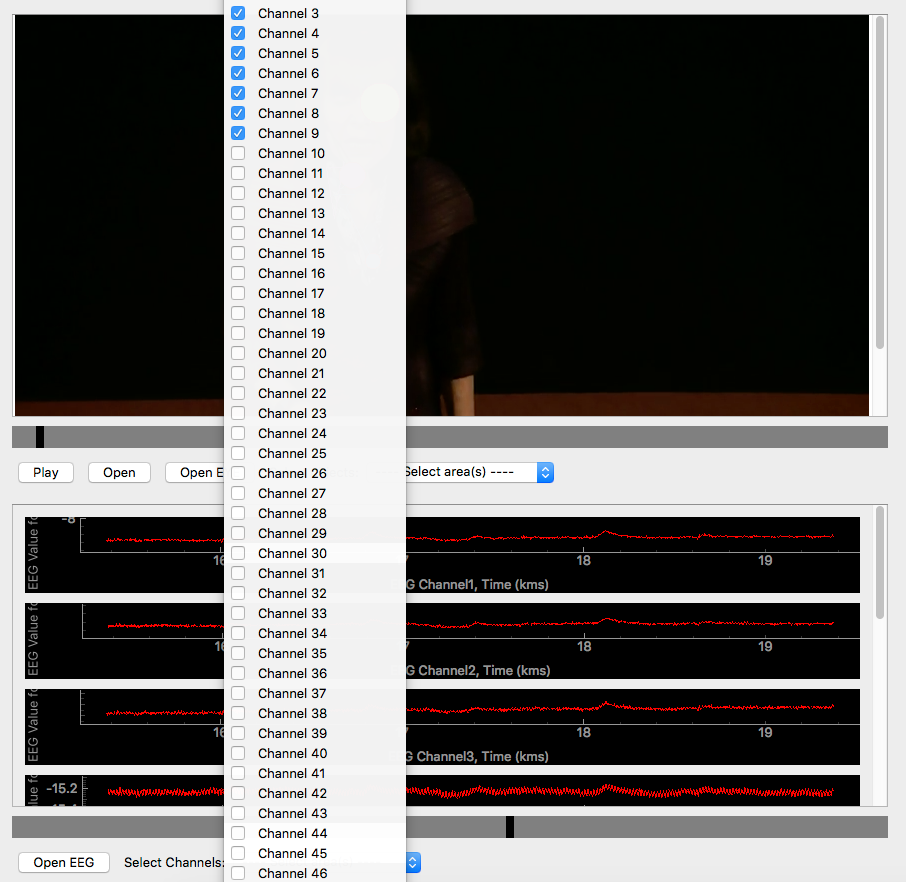}}\\
   \caption{(a) is the display when selecting audiences' eye data. (b) is the display when selecting the channels}
   \label{fig:toolselection}
\end{figure}

In this section, we show several screen shots for all functionalities and views. To better present the eye trail lines, we select a moment when viewers are reading the slides. Similarly, we choose a moment the viewers are gazing at the speaker to show the fixation. Two screen shots are illustrated in Fig.\ref{fig:toolnormal}
We also show the combobox when user selecting the audience's eye movement and the channels of EEG in Fig.\ref{fig:toolselection}

\section{Conclusions and Future Works}

We have compared different methods to calculate the similarity of eye movement between pairs of viewers. After studied several approaches including mechanical statistics, etc., we decide to use time series matching methods. Adopting TWED method, we are able to get similarity matrix for any specified time window. However, due to the fact we don't have reliable attention ground-truth, the hyper-parameter selection and the assessment of the similarity matrix remains a challenge. We also try to use question answer correctness as attention ground-truth but also find it's unreliable due to lack of enough questions. However, the visual assessment by showing eye movement path indicates promising trend. By using the similarity matrix, we are able to cluster the audience by using community detection algorithm. Finding the attention ground-truth seems to be a critical task for future study. By tackle this problem, we would be able to better select the model, hyper-parameters and assess the similarity matrix and clustering result. Another work we have done is developing the visualization tool. Although the functionality is relatively basic, but after exploring the suitable architecture and pattern, we would spent much less effort when adding clustering visualization or other functionalities in the future.

\bibliographystyle{splncs}
\bibliography{egbib}

\begin{subappendices}
\renewcommand{\thesection}{\Alph{section}}%

\section{Output Format of Eye-tracking data}
The files are located in \\ \texttt{nsf\char`_code\char`_and\char`_data\char`_package\char`_2017\char`_06\char`_25}$\backslash$\texttt{nsf\char`_ted\char`_matlab\char`_data} \\ 
There two kinds of files came out of the eye-tracking machine: \\
\begin{description}
  \item[$\bullet$] 1.The \textbf{asc files}, which contains the raw data out of the machine.\\ Each audience for a lecture would have his or her own file. \\
The file's name usually is arranged as \texttt{\{objname\}\char`_\{lecturename\}.asc}, for example, \texttt{msb\char`_c.asc}. \\
The file could be opened as simple txt file. Although this kind of file contains all the raw information, most of the useful information inside would also be included in the xls file with better aggregation. The most useful information to us in the file is MSG 0 \texttt{DISPLAY\char`_COORDS 0 0 \{WIDTH\} \{HEIGHT\}}. For example, 0 \texttt{DISPLAY\char`_COORDS} 0 0 1023 767. 
  \item[$\bullet$] 2. The \textbf{xls files}, which contains the aggregated data from the raw data. \\ Each audience for a lecture would have his or her own file. \\
  The file's name usually is arranged as \texttt{\{objname\}\char`_\{lecturename\}.xls}, for example, \texttt{an1\char`_carol.xls}. The file could be opened as a simple excel file.\\
The file contains 222 columns. Many of them containing event data such as \texttt{CURRENT\char`_FIX\char`_BUTTON\char`_0\char`_PRESS}, which in most case is empty. Among with columns, \texttt{CURRENT\char`_FIX\char`_DURATION, CURRENT\char`_FIX\char`_START}, \texttt{CURRENT\char`_FIX\char`_END}, \texttt{CURRENT\char`_FIX\char`_X and CURRENT\char`_FIX\char`_Y} are most important to us. The unit of the columns \texttt{CURRENT\char`_FIX\char`_DURATION, CURRENT\char`_FIX\char`_START} and \\
\texttt{CURRENT\char`_FIX\char`_END} are all millisecond. The unit of the position columns \\ \texttt{CURRENT\char`_FIX\char`_X} and \texttt{CURRENT\char`_FIX\char`_Y} are the pixel location relative to the left up corner of the screen.
\end{description}

\section{TWED Algorithm in our application}

\noindent
{\it Modified iterative implementation of the TWED distance}
\begin{verbatim}
float TWED(t1_data[1 to n], t2_data[1 to m], lam, nu):
	int result(result)
	result = init_matrix(result)
    for i := 1 to m
 		result[0,i] := infinity;
 	for i := 1 to n
 		result[i,0] := infinity;
 		result[0,0] := 0; 
    n = len(t1_data)
    m = len(t2_data)
    for p = 1 to n:
        for q = 1 to m:
            insertion = result[p - 1][q] 
            				+ Dist(t1_data[p - 1], t1_data[p]) + nu  + lam
            deletion = result[i][q - 1] 
            				+ Dist(t2_data[q - 1], t2_data[q]) + nu + lam
            match = result[p - 1][q - 1] 
            			+ Dist(t1_data[p], t2_data[q]) 
                     + 2 * nu * (abs(p - q)) 
                     + Dist(t1_data[p - 1], t2_data[q - 1])
            result[p][q] = min(insertion, deletion, match)
    return result[n - 1][m - 1]
\end{verbatim}
\noindent
{\small code modified based on "Iterative implementation of the TWED distance" in Marteau et al.\cite{marteau2009time}}
\clearpage
\section{TWED Hyperparameter Setting Comparison}
\begin{figure}[htbp]
   \centering
   \subfloat[][]{\includegraphics[width=.47\textwidth]{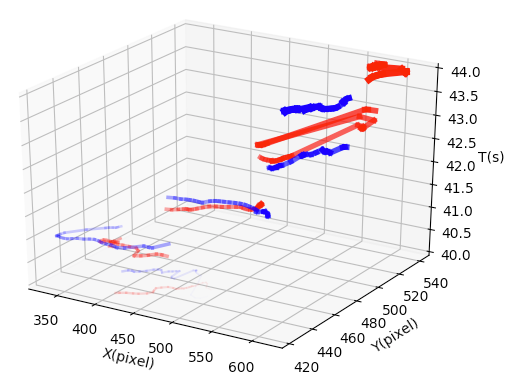}}\quad
   \subfloat[][]{\includegraphics[width=.47\textwidth]{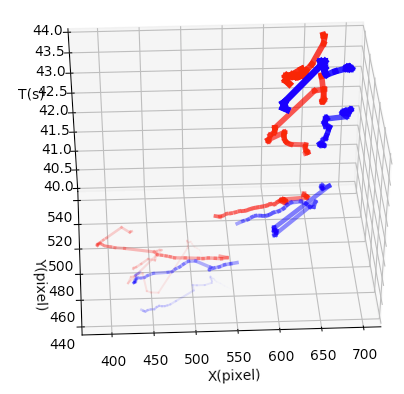}}\\
   \subfloat[][]{\includegraphics[width=.47\textwidth]{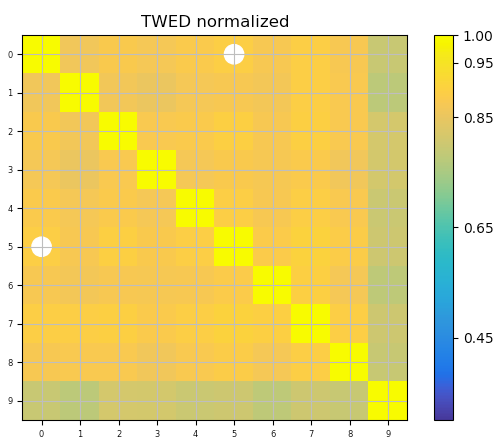}}\quad
   \subfloat[][]{\includegraphics[width=.47\textwidth]{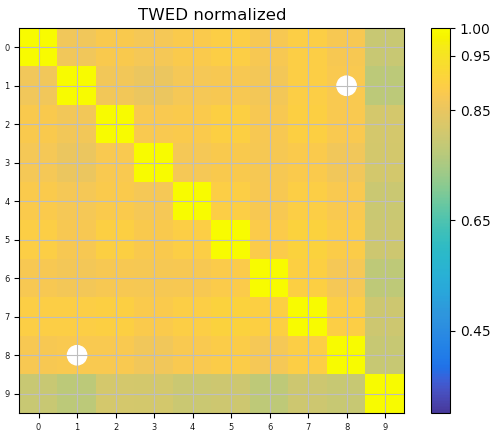}}\\
\end{figure}  
\begin{figure}[htbp]\ContinuedFloat
   \centering
   \subfloat[][]{\includegraphics[width=.47\textwidth]{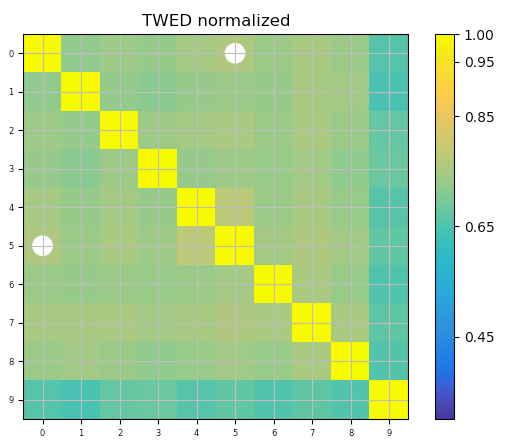}}\quad
   \subfloat[][]{\includegraphics[width=.47\textwidth]{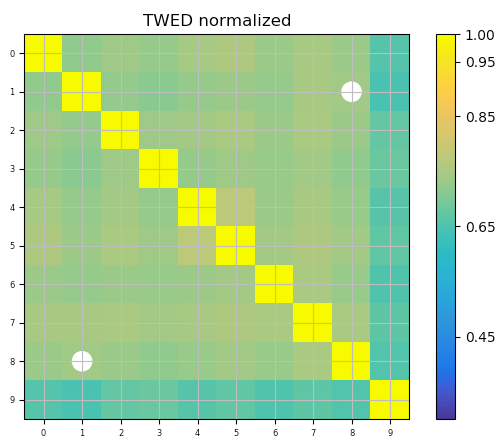}}\\
   \subfloat[][]{\includegraphics[width=.47\textwidth]{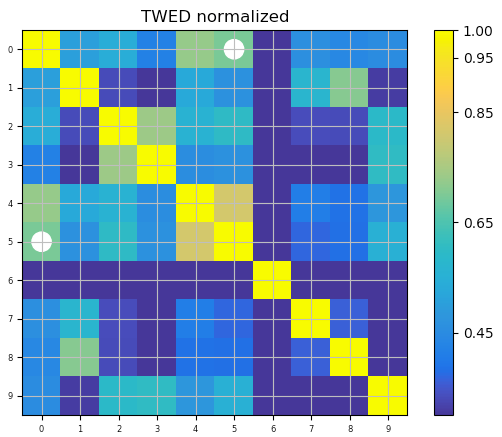}}\quad
   \subfloat[][]{\includegraphics[width=.47\textwidth]{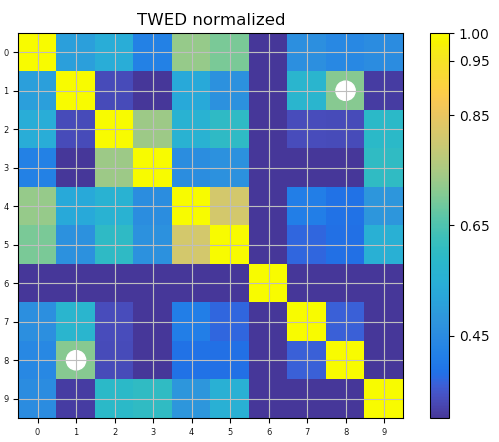}}\\
   \caption{(a) shows the eye trails of audience 0 and 5 during 45s-50s in TED Talk Carol,(b) shows the eye trails of audience 1 and 8 during 45s-50s in TED Talk Carol. (c) and (d) are the similarity matrix under corresponding setting when having $\lambda = 0$ and $\gamma = 0$.(e) and (f) are the similarity matrix under corresponding setting when having $\lambda = 5000$ and $\gamma = 5000$.(g) and (h) are the similarity matrix under corresponding setting when having $\lambda = 10000$ and $\gamma = 100000$}
\label{fig:comp}
\end{figure}

\clearpage
\section{Fast Multi-Scale Community Detection Algorithm}
\noindent
\begin{verbatim}
Initialise current community partition with a node per community:
                                          com = list of all nodes
for all scale parameters p do
    Compute initial Q value given p: Q = computeQ(com, p)
    while changes can be made do
        while nodes can be moved do
            nlist = list of all nodes
            while nlist is not empty do
                n = pick a random node in nlist
                ncom = neighbour communities of n
                best_\deltaQ = 0
                for all communities nc in ncom do
                     Compute the \delta Q that moving n into nc 
                                                  would produce
                     if \deltaQ > best_\deltaQ and move does 
                                  not break a community then
                         best_\deltaQ = \delta Q
                         best_c = nc
                     end if
                 end for
                 if best_\delta Q > 0 then
                     Update com: move node n to community best_c
                     Update total value of Q: Q = Q + best_\delta Q
                 end if
            end while
        end while
        while clusters can be merged do
            clist = list of all current communities
                while clist is not empty do
                     c = pick a random community in clist
                     ncom = neighbour communities of c
                     best_\delta Q = 0
                     for all communities nc in ncom do
                        	Compute the \delta Q that merging c and nc
                        	                             would produce
                        if \deltaQ > best_\deltaQ then
                             best_\delta Q = \delta Q
                             best_c = nc
                        end if
                     end for
                     if best_\deltaQ > 0 then
                         Update com: merge communities c and best_c
                         Update total value of Q: Q = Q 
                                               + best_\deltaQ
                     end if
                end while
        end while
    end while
    Store com and Q for p
end for
return Community sets and associated Qs
\end{verbatim}
\noindent

\section{Dependencies of Development}

\begin{table}
\begin{center}
\caption{These are the packages that should be installed in order to develop. Using anaconda package-manager is recommended.}
\label{table:headings}
\begin{tabular}{lll}
\hline\noalign{\smallskip}
Package name & Version & Notes\\
\noalign{\smallskip}
\hline
Python  & {2.7} & 2.6 or 3.4 no guarantee \\
PyQt5  & {5.6.0} & anaconda support up to 5.6.0, higher version should be compatible\\
pyqtgraph  & {0.10.0} & anaconda supported\\
numpy  & {1.13.3} & anaconda supported, higher version should be compatible\\
pandas & {0.20.3} & anaconda supported, higher version should be compatible\\
\hline
\end{tabular}
\end{center}
\end{table}
\setlength{\tabcolsep}{1.4pt}

\section{GUI Operation}

\begin{table}
\begin{center}
\caption{GUI element and their triggered functions}
\label{table:headings}
\begin{tabular}{|p{2cm}|l|p{8cm}|}
\hline\noalign{\smallskip}
GUI Element & Event & Description\\
\noalign{\smallskip}
\hline
Open  & click & Load the new video files, multiple format supported \\ & & \\
Open EYE  & Click & Pop up a multi-files select window. load the eye tracking data and preprocess them. The file should be csv or excel, sample files are located under project repo's fixation folder \\ & & \\
Open EEG  & Click & Pop up a  single file select window. load the EEG tracking data and preprocess them. The file should be txt, sample files are the  \texttt{AZZ1\char`_v2\char`_Carol\char`_30sec.txt} inside the project repo\\ & & \\
Select Objects  & Click & Pop up the audiences selection combobox \\ & & \\
Select Objects Combobox & Check options & Add elements of the checked audience or removed the elements of the unchecked audience \\ & & \\
Select Channels & Click & Pop up the channels selection combobox \\ & & \\
Select Channels Combobox & Check options & Add plot of the checked channel or removed the plot of the unchecked Channels \\ & & \\
Play & Click & Video starts to play if any video uploaded. The button's label will change to Pause \\ & & \\
Pause & Click & Video stops if any video uploaded. The button's label will change to Play \\ & & \\
Video Slider & Drag & Video stops and the video lapse would change to the corresponding position.\\ & & \\
Video Slider & Auto Update & Video, Eye movement elements and EEG would be all on sync with the slider's position.\\ & & \\
\hline
\end{tabular}
\end{center}
\end{table}
\setlength{\tabcolsep}{1.4pt}

\end{subappendices}

\end{document}